\def\year{2019}\relax
\documentclass[letterpaper]{article} 
\usepackage{aaai19}  
\usepackage{times}  
\usepackage{helvet}  
\usepackage{courier}  
\usepackage{url}  
\usepackage{graphicx}  
\usepackage{algorithm}
\usepackage{amsmath}
\usepackage{amssymb}
\usepackage[noend]{algpseudocode}
\usepackage{subcaption}
\usepackage {tikz}
\usetikzlibrary {positioning}

\newcommand{\pluseq}{\mathrel{+}=}

\begin{document}
%
\title{Generation of Policy-Level Explanations for Reinforcement Learning}
\author{
Nicholay Topin \ and Manuela Veloso \\
Machine Learning Department\\
Carnegie Mellon University\\
Pittsburgh, PA 15213 \\
\texttt{\{ntopin, veloso\}@cs.cmu.edu} \\
}
\maketitle
\begin{abstract}
Though reinforcement learning has greatly benefited from the incorporation of neural networks, the inability to verify the correctness of such systems limits their use. 
Current work in explainable deep learning focuses on explaining only a single decision in terms of input features, making it unsuitable for explaining a sequence of decisions. 
To address this need, we introduce Abstracted Policy Graphs, which are Markov chains of abstract states. 
This representation concisely summarizes a policy so that individual decisions can be explained in the context of expected future transitions. 
Additionally, we propose a method to generate these Abstracted Policy Graphs for deterministic policies given a learned value function and a set of observed transitions, potentially off-policy transitions used during training. 
Since no restrictions are placed on how the value function is generated, our method is compatible with many existing reinforcement learning methods. 
We prove that the worst-case time complexity of our method is quadratic in the number of features and linear in the number of provided transitions, $O(|F|^2 |tr\_samples|)$. 
By applying our method to a family of domains, we show that our method scales well in practice and produces Abstracted Policy Graphs which reliably capture relationships within these domains.
\end{abstract}

\section{Introduction}
Recent advances in neural networks have led to powerful function approximators, which have been successfully used to support Reinforcement Learning (RL) techniques to solve difficult problems.
However, the deployment of RL systems in real-world use cases is hampered by the difficulty to verify and predict the behavior of RL agents. 
In the context of RL, autonomous agents learn to operate in an environment through repeated interaction. 
After training, the agent is able to make decisions in any given state, but is unable to provide a plan nor rule-based system for determining which action to take. 
Generally, a policy  which selects actions ($\pi(s) = a$) is available along with its value function ($V_{\pi}(s) \in \mathbb{R}$), which predicts future reward from a state. 
However, neither the outcome of the actions nor the sequence of future actions taken is available. 
Without these, a human operator must blindly trust an RL agent's evaluation. 

Existing techniques for explaining Deep Reinforcement Learning agents borrow techniques used for explaining neural network predictions, so they focus on explaining one state at a time. 
These techniques pinpoint the features of the state that influence the agent's decision, but do not provide an explanation incorporating expected future actions.
Therefore, the explanation is insufficient for a human supervisor to decide whether to trust the system. 
Likewise, no whole-policy view is available, so evaluating the agent's overall competency (as opposed to single-state evaluation) is impossible. 
For these reasons, we are interested in explaining policies as a whole: giving the context for action explanations and providing an abstraction of an entire policy. 

To address the aforementioned issues, we propose the creation of a full-policy abstraction, which is then used as the basis for generating local explanations.
We introduce Abstract Policy Graphs (APGs) as such a full-policy abstraction. 
Each APG is effectively a graph where each node is an abstract state and each edge is an action with associated transition probability between two abstract states.
Using a mapping from states to abstract states, one can identify which groups of states the agent treats similarly, as well as predict the sequence of actions the agent will take. 
This explanation provides local explanations along with a global context. 

Additionally, we propose an algorithm, APG Gen, for creating an APG given a policy, a learned value function, and a set of transitions. 
Starting with a single abstract state which encompasses the full state-space, APG Gen uses a feature importance measure to repeatedly divide abstract states along important features. 
These abstract states are then used to create an APG. 
The splitting procedure additionally identifies which features are important within each abstract state. 
Notably, this general procedure is compatible with existing methods for learning a policy and value function.

The main contributions of this work are as follows: (1) we introduce a novel representation, Abstract Policy Graphs, for summarizing policies to enable explanations of individual decisions in the context of future transitions, (2) we propose a process, APG Gen, for creating an APG from a policy and learned value function, (3) we prove that APG Gen's runtime is favorable ($O(|F|^2 |tr\_samples|)$, where $F$ is the set of features and $tr\_samples$ is the set of provided transitions), 
and (4) we empirically evaluate APG Gen's capability to create the desired explanations.

\section{Related Work}
Prior work in explaining Deep Learning systems focuses on explaining individual predictions. 
Explaining individual predictions in terms of input pixels has been done using saliency maps based on model gradients, as in \cite{image_FI_13} and \cite{image_FI_samek}. 
Alternatively, local explanations are learned for regions around an input point to identify relevant pixels \cite{image_FI_dog}. 

\cite{katia} leverage these pixel-level explanations and use an object detector to produce object saliency maps, which are explanations for the behavior of deep RL agents in terms of objects. 
Unlike our method, their method explains a single decision without the context of potential future decisions. 
\cite{frogger} create natural language explanations for each action the agent performs. 
These explanations are learned using a human-provided corpus of explanations. 
These explanations are only for individual $(s,a,s')$ tuples, so their method does not produce policy-wide explanations.

Existing RL-specific methods with policy-level explanations impose additional constraints. 
Some works explain agent behavior, but require the agent to use a specific, interpretable model. 
For example, Genetic Programming for Reinforcement Learning \cite{genetic} use a genetic algorithm to learn a policy which is inherently explainable. 
Unlike our method, this method is incompatible with arbitrary RL systems due to its reliance on learning inherently small policies using a genetic algorithm. 

Other methods require a ground-truth or learned model of the environment, which may be more complicated than the learned policy. \cite{MinSufExp} produce contrastive explanations which compare the agent's action to a proposed alternative, but require a known factored MDP. \cite{ImpRoboticController} explain robot behavior using natural language. These explanations are formed using a model learned from demonstrations and based on operator-specified ``important program state variables'' and ``important functions.'' 
Other work avoids automatically identifying patterns in agent behavior and relies on a human to manually identify similar sets of states. \cite{cluster_drl} embed states into a space where states within certain regions of this space behave similarly. They group these states based on region to produce explanations. However, a human operator must form these groups and identify within-group similarities. 

An overview of previous methods for creating abstractions for Markov Decision Processes can be found in \cite{UnifiedAbstraction}. These methods focus on creating an abstraction for use with an RL agent. To that end, they create an abstract Markov Decision Process, which is usually done before or during learning. 
This differs from our use case where we seek only to explain the transitions that occur under a specific policy, so our abstraction is instead a Markov chain created after a policy has been learned.

\section{Background}
\subsection{Solving Markov Decision Processes} \label{background_MDPs}
In the context of RL, an agent acts in an environment defined by a Markov decision process (MDP). 
We use a six-tuple MDP formulation: $\langle S, A, P, R, \gamma, T \rangle$, where $S$ is the set of states, $A$ is the set of actions, $P$ is the transition function, $R$ is the reward function, $\gamma$ is the discount factor, and $T$ is the set of terminal states which may be the empty set \cite{SuttonAndBarto}. 
In this work, we assume all states consist of an assignment to features. 
Specifically, each state consists of a value assignment to each feature $f \in F$. 
An agent ultimately seeks to learn a policy, the function $\pi(s_t) = a_t$, which maximizes total discounted reward.
Note that the policy need not be deterministic, but we only consider the deterministic case in this paper. 
In the process of learning a policy, an RL agent generally approximates the state-value function or the action-value function. 
The state-value function is the expected future discounted reward from the state $s_0$ when policy $\pi$ is followed:
\begin{equation} \label{value}
V_{\pi}(s_0) = \mathbb{E} \left( \sum_{t=0}^{\infty} \gamma^t R(s_t, \pi(s_t), s_{t+1} ) \right). 
\end{equation}
The action-value function is the expected future discounted reward from the state $s_0$ given the agent takes action $a_0$ and then follows policy $\pi$.
Note that the state-value function can be obtained from the action-value function: $V_{\pi}(s_0) = Q_{\pi}(s_0, \pi(s_0))$. 
These are used in methods based on Q-Learning, Sarsa($\lambda$), and actor-critic methods \cite{SuttonAndBarto}. 
Therefore, the value-function is generally available alongside the policy of a trained agent. 

\subsection{Feature Importance Function} \label{approach_featureImportanceFunction}
We use an \emph{importance measure} for grouping states from an original MDP into abstract states. 
An importance measure is a function $I_f(c)$ which represents the \emph{importance} of feature $f$ in determining how a system treats a set of inputs (e.g., states), $c$. 
If $f$ takes on the same value for all $s \in c$ or its value does not influence the system's output, then $f$ is not important. 
We use the Feature Importance Ranking Measure (FIRM) \cite{FIRM} since it is fast to compute exactly for binary features and can be meaningfully interpreted. 

To calculate importance, FIRM uses $q_f(v)$, the \textit{conditional expected score} of $s$ for a feature $f$ with respect to an arbitrary function $g(s)$. 
This score is the  average value of $g(s)$ for all $s$ within the set $c$ where feature $f$ takes value $v$: 
\begin{equation}
q_f(v) = \mathbb{E}(g(s) | s[f] = v).
\end{equation}

Intuitively, if $q_f(v)$ is a flat function, then $v$, the value of $f$, has no impact on the average value of $g(s)$ over $s \in c$, so provides little information. 
However, if $v$ significantly impacts $g(s)$, then the value of $q_f(v)$ will vary. 
This motivates FIRM's importance measure $I_f(c)$, the variance of the conditional expected score:
\begin{equation}
I_f(c) = \sqrt{\mathbb{V}(q_f(s[f]))}.
\end{equation}

In specific cases, the exact value of $I_f(c)$ can be computed quickly. One such case is when $f$ is a binary feature. 
For a binary feature, the importance measure is given by 
\begin{equation}
\begin{split}
I_f(c) = (q_{f0}(c) - q_{f1}(c)) \sqrt{ p_{f0}(c) p_{f1}(c)},\\
p_{fv}(c) = \mathbb{P}(s[f] = v),\\
q_{fv}(c) = \mathbb{E}(g(s) | s[f] = v).
\end{split}
\end{equation}
In the case of binary features, FIRM corresponds to the expected change if the feature switches from 0 to 1. 
Conveniently, sign is preserved in the binary case, showing magnitude of importance as well as direction of effect.

\section{Approach}
In Section~\ref{approach_graphRepresentation}, we describe Abstract Policy Graphs, our representation for explaining a policy. 
In Section~\ref{approach_graphConstruction}, we propose APG Gen, a method for constructing such explanations. 
In Section~\ref{approach_abstractStateSummarization}, we describe local explanations we produce from our policy explanations. Finally, in Section~\ref{approach_computationalComplexity}, we show that our method has favorable asymptotic runtime: quadratic in the number of features and linear in the number of transition tuples considered, where there are usually few features and runtime sub-linear in the number of transitions is unattainable. 

\subsection{Abstract Policy Graphs} \label{approach_graphRepresentation}
\begin{figure}[t]
\includegraphics[width=\linewidth]{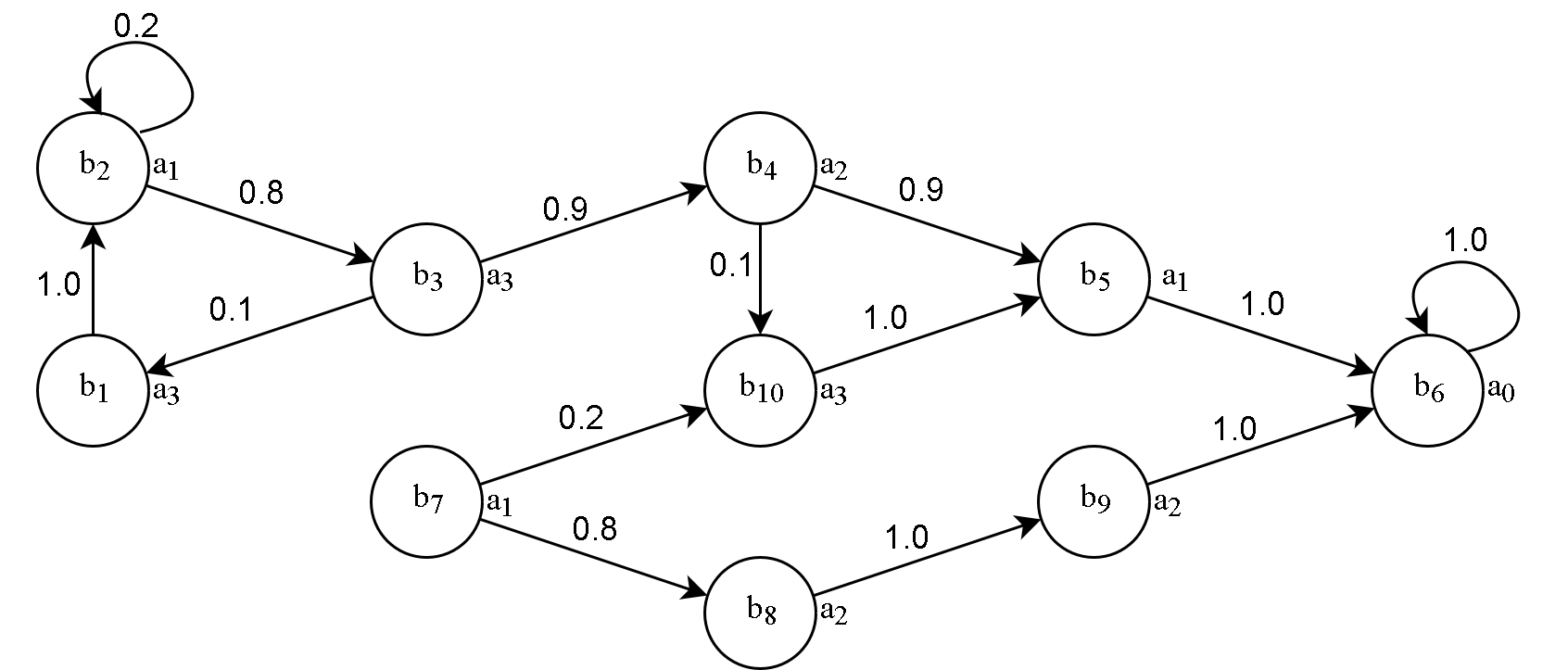}
\caption{An example Abstract Policy Graph with edge labels indicating transition probabilities. The abstract state identifier is shown within each node, and the action taken is written adjacent to the node.}
\label{APG_example}
\end{figure}

To create a policy-level explanation, we express the policy as a Markov chain over abstract states where edges are transitions induced by a single action from the original MDP, which we term an Abstract Policy Graph (APG). We present an example in Figure~\ref{APG_example}.
Consider a mapping function, $l(s)$, which maps states in the original MDP ($grounded$ states) to abstract states. In effect, each abstract state represents a set of grounded states from the original MDP. 
We use the phrase ``an agent is \textit{in abstract state $b$}'' to mean that the current state of the domain, $s$, maps to $b$ (i.e., $l(s) = b$).
Let each set contain all states \textit{interchangeable} under the agent's policy such that the agent behaves similarly when starting in a state from the set in a similar fashion. As a result, states in which the agent behaves similarly lead the agent to states in which the agent also behaves similarly. If the agent's transitions between grounded states are approximated using a Markov chain between these abstract states, then states which are treated similarly are readily identified and the agent's transitions between abstract states can be predicted.

For example, let the agent's distribution of actions be approximately equal for all future time-steps for all grounded states in the set: 
\begin{equation} \label{inter}
\mathbb{E}_{s_{1,t}} \mathbb{P}(\pi(s_{1,t}) = a) \approx \mathbb{E}_{s_{2,t}} \mathbb{P}(\pi(s_{2,t}) = a) \forall a, t
\end{equation}
for all $s_1$ and $s_2$ within a set, where $s_{i,t}$ is the state reached from $s_{i}$ in $t$ further time-steps using policy $\pi$.

If the transition function is deterministic, then the agent takes the same sequence of actions from each grounded state in the set because there is only a single $s_{i,t}$ for each $i$ and $t$. In addition, since no two sets could combine to form an interchangeable set, then the abstract states for all $s_{i,t}$ are identical, too. Since the probability of transitioning from one abstract state to another after one action is then either zero or one (regardless of grounded state), the agent is effectively traversing a Markov chain of abstract states induced by its policy. 

However, most interesting domains have stochastic transition functions. 
Under a stochastic transition function, two grounded states can satisfy Equation~\ref{inter} while having different transition probabilities to future abstract states. The transition probability from one abstract state to another can be approximated as the average transition probability for grounded states in the source abstract state. For stochastic policies, the probability of taking any given action can be similarly approximated as the average over all grounded states in the source abstract state. The transition probability is no longer exact for any given grounded state in the set but is the transition probability for a randomly chosen state in the set. 
To make predictions for a series of transitions, we make a simplifying Markov assumption:
the abstract state reached, $b_{t+1}$, when performing an action depends only on the current abstract state, $b_{t}$. This assumption leads to approximation error but works well in practice, as shown in Section~\ref{ex_markov_nhop}.

The abstract states now form a Markov chain, as desired. This final product allows human examination of higher-level behavior (e.g., looking at often-used trajectories and checking for loops), prediction of future trajectory (along with accompanying probability), and verification of agent abstraction (e.g., ensuring agent's behavior is invariant to certain features being changed).

\subsection{APG Construction} \label{approach_graphConstruction}
We propose an algorithm for creating APGs, APG Gen. It first divides states into sets to form abstract states, then computes transition probabilities between them. 

\subsubsection{Importance Measure}
APG Gen is compatible with arbitrary interchangeability measures. We choose $V_{\pi}(s)$ as it is readily available, but the method does not rely on this choice. Using an interchangeability measure based on Equation~\ref{inter} is difficult since it would depend on an expectation over all future states, which is often computationally expensive and requires knowing transitions for all states. We note that the definition of $V_{\pi}(s)$ in Equation~\ref{value} also includes an expectation over all future future states, as well as a dependency on the policy. Since the full state-value function is generally available and does not require computing additional expectations, we use it as our measure of interchangeability. The intuition is that two states with similar state-values lead to similar future outcomes in terms of reward, so are likely treated similarly by the agent. A different measure could be used instead

With an importance measure $I_f(c)$ for $V_{\pi}(s)$, a set of states which is interchangeable under the agent's policy should have low $I_f(c)$ for all $f$. Consider the case of $c_1 \bigcup c_2$, a set containing the original MDP states which should be contained in two abstract states. 
At least one $f$ should have high $I_f(c_1 \bigcup c_2)$ because the grounded states from the two abstract states are treated differently. 
If there is no such $f$, then the two sets are treated the same and therefore belong to the same abstract state.

\subsubsection{Splitting Binary Features}
In the case where all features are binary, if the set is split based on the value of that $f$ (into one subset if $f=0$ and the other if $f=1$), then both subsets will have $I_f$ be 0, since $f$ has a constant value within the subset. This holds for any set of grounded states which will ultimately form several abstract states. Therefore, this splitting procedure can be repeatedly performed to create abstract states from initially larger sets until all features have low importance.

Since each binary feature can only be important once and it is straightforward to split along a binary feature, the use of binary features allows quick computation. Therefore, in cases where the original MDP does not have solely binary features, pre-processing can be done to create features for APG Gen. Note that these features are not used when evaluating $V_{\pi}(s)$ (i.e., an unmodified, arbitrary model can be used for approximating $V_{\pi}(s)$). The binary features are instead used when deciding to which set a specific tuple belongs while performing APG Gen. 

\subsubsection{Abstract State Division}
Since binary features allow efficient splitting, our approach is to initially form sets based on action taken under the current policy and then repeatedly split the set which has the greatest $I_f$ value, as computed within that set. When the importances of all features for all abstract states are sufficiently low, then the abstract states consist of sets of states which are interchangeable under the agent's policy. 

The pseudocode for our method is given in Algorithm~\ref{alg_makePolicyGraph}. To perform the procedure, we require a set of sample transitions. As mentioned in Section~\ref{background_MDPs}, RL agents generally learn through interacting with a domain, meaning a set of $(s,a,s')$ transitions is generally available. The notation we use for this set is a vector $tr\_samples$ consisting of entries $t$ where action $t_a$ is taken in state $t_s$, leading to a transition to state $t_{s'}$, an observed reward $t_r$, and a termination flag $t_t$ (0 or 1). The policy is used in line 5 to discard transition tuples where the provided policy would perform a different action from the action in the stored tuple. This is done so that the generated explanation reflects only the current policy and not transition tuples observed under past policies.

Lines 2-6 separate the tuples based on the action taken. We pre-compute the feature importance for each set and save it in lines 7-8. Line 9 forms the core procedure, where abstract states are divided until no feature has importance greater than $\epsilon$. The abstract state with the most important feature is found, then divided based on the most important feature. The importance of each feature is then re-computed in lines 18 and 20.

\begin{algorithm}[t] 
\caption{Compute abstract states based on transition samples and learned policy.} \label{alg_makePolicyGraph}
\begin{algorithmic}[1]
  \Procedure{Div\_Abs\_States}{$tr\_samples, policy$} 
    \For {$i \textrm{ in } \{1, \ldots, |A| \}$}
      \State $c[i] \leftarrow \varnothing$ \Comment{initially, all sets empty}
    \EndFor
    \For {$t \textrm{ in } tr\_samples$} \Comment{separate by action}
      \If {$policy(t_s) = t_a$} 
        \State $c[t_a] \leftarrow c[t_a] \cup t$ 
      \EndIf
    \EndFor
    \For {$i \textrm{ in } \{1, \ldots, c\}$} \Comment{pre-compute feat. imp.}
      \State $m[i] \leftarrow [ | I_f(c[i]) | \textbf{ for } f \in \{1, \ldots, |F|\}]$ 
    \EndFor
    \While {$\text{max}_i \text{max}_j \left( m[i][j] \right) > \epsilon$}
      \State $i_{max} \leftarrow \text{argmax}_i \text{max}_j \left( m[i][j] \right)$
      \State $j_{max} \leftarrow \text{argmax}_j \left( m[i_{max}][j] \right)$ 
      \State $c_{n_0}, c_{n_1} \leftarrow \varnothing$
      \For {$t \textrm{ in } c[i_{max}]$} \Comment{split on most imp. feat.}
        \If{$t_s[j_{max}] = 0$}
          \State $c_{n_0} \leftarrow c_{n_0} \cup t$
        \Else
          \State $c_{n_1} \leftarrow c_{n_1} \cup t$
        \EndIf
      \EndFor
      \State $m[i_{max}] \leftarrow [|I_f(c_{n_0})| \textbf{ for } f \in \{1, \ldots, |F|\}]$ 
      \State $c[i_{max}] \leftarrow c_{n_0}$
      \State $m[|c|] \leftarrow [|I_f(c_{n_1})| \textbf{ for } f \in \{1, \ldots, |F|\}]$
      \State $c[|c| + 1] \leftarrow c_{n_1}$
    \EndWhile
    \State $\textbf{return } c$
  \EndProcedure
\end{algorithmic}
\end{algorithm}

\subsubsection{APG Edge Creation}
Once the abstract state sets have been created, we create the mapping function $l$ and Markov chain transition matrix using Algorithm~\ref{alg_computeTransitions} (a sparse matrix can be created in an almost identical fashion). In lines 2-4, the contents of each set are used to create the necessary entries in a lookup table for the mapping function. Simultaneously, the transition matrix is initialized to be zero-valued by lines 5-6. Then, in lines 7-13, the mapping function is used in conjunction with the transition tuples within each abstract state set to compute transition probabilities. That is, if a transition  tuple $t$ is in set $c[i]$, then the origin state, $t_s$, is in the abstract state represented by $c[i]$. The destination state, $t_{s'}$, then indicates a connection between $c[i]$ and $l(t_{s'})$. The transition probability from $c[i]$ to $c[n]$ is the portion of tuples in $c[i]$ which lead to a state in $c[n]$, so each tuple in $c[i]$ should increment $transition(i, l(t_{s'}))$ by $1 / |c[i]|$, as done in line 13. 
Terminal transitions are identified in line 9 and instead lead to the special $b_T$ abstract state. This abstract state represents termination and is represented by the highest-numbered row and column in $transition$. There will be no incoming edges to $b_T$ if the set of terminal states, $T$, is empty. Line 14 sets $b_T$ to have an edge to itself to create a valid Markov chain.

\begin{algorithm}[t] 
\caption{Create mapping function and transition matrix based on policy graph.} \label{alg_computeTransitions}
\begin{algorithmic}[1]
  \Procedure{Compute\_Graph\_Info}{$c$} 
    \For {$i \textrm{ in } \{1, \ldots, |c|+1\}$} 
      \For {$t \textrm{ in } c[i]$}
         \State $lookup[t_s] \leftarrow i$ \Comment{create lookup table}
      \EndFor
      \For {$n \textrm{ in } \{1, \ldots, |c|+1\}$} \Comment{zero matrix}
        \State $transition(i, n) \leftarrow 0$ 
      \EndFor
    \EndFor
    \For {$i \textrm{ in } \{1, \ldots, |c|\}$}
      \For {$t \textrm{ in } c[i]$}
        \If{$t_t = 1$} \Comment{terminal $t_t$ go to dummy $b_T$}
          \State $n \leftarrow |c| + 1$
        \Else \Comment{others go to abstract state of next state}
          \State $n \leftarrow lookup[t_{s'}]$
        \EndIf
        \State $transition(i, n) \pluseq 1 / |c[i]|$
      \EndFor
    \EndFor
    \State $transition(|c|+1, |c|+1) \leftarrow 1$ \Comment{add $b_T$ self-loop}
    \State $\textbf{return } lookup, transition$
  \EndProcedure
\end{algorithmic}
\end{algorithm}

\subsection{Abstract State Summarization} \label{approach_abstractStateSummarization}
The policy graph algorithm presented creates a summary of the overall policy out of abstract states, which are each defined by a set of states from the original MDP. Due to the process which we use to create the abstract states, we can also create a characterization of the states which are in their set. Note that Algorithm~\ref{alg_makePolicyGraph} splits an abstract state into two based on a feature $f$ because $f$ is ``important'' based on chosen function $g$. These $f$s can be trivially recorded and stored for each abstract state. Once the final abstract states have been created, these $f$s indicate which features were previously important. From this, the important features of an abstract state can be determined. 

For any state in the transition set, $s_n$, and a specific abstract state, $b$, if $\pi(s_n) = \pi(s)$ and $s_n[f] = s[f]$ for any $s$ in $b$'s set and for all $f$ which were used to create $b$, then $s_n$ will also be in $b$'s set. Similarly, if $s_n[f] \neq s[f]$ (for similarly defined $s$ and $f$), then $s_n$ cannot be in $b$'s set. These feature-value assignments are necessary and sufficient to be part of $b$, so this creates an ``if and only if'' relationship. As a result, for any chosen state $s$, based on the features used to create its abstract state, the ``relevant'' features can be determined. 
If the value for any of these features changes, then $s$ would be in a different abstract state and treated differently. Similarly, the agent is oblivious to changes in the other features given the values assigned to the relevant features. This relationship allows a human supervisor to determine which features affect how an agent treats a specific state. In addition, a summary of an abstract state can be formed using these same feature-value assignments. 

\subsection{Computational Complexity} 
\label{approach_computationalComplexity}

\subsubsection{Computing FIRM}
Since $p_{f1}(c) = 1 - p_{f0}(c)$ and $q_{f1}(c) = (\mathbb{E}(g(s)) - p_{f0}(c) q_{f0}(c)) / p_{f1}(c)$, computing $p_{f0}(c)$ and $q_{f0}(c)$ is enough to calculate the importance. These can be computed for all $f$ with a single pass through the set of states, as shown in Algorithm~\ref{alg_firm}.

The bulk of the computation is performed in lines 6 to 12. Here, every transition in the set is separately considered. Only a single evaluation of $g$ is required regardless of the number of features. This evaluation is used to calculate $\mathbb{E}(g(s))$ and $q_{f0}$ for each feature where $s[f] = 0$. The overall complexity of Algorithm~\ref{alg_firm} is therefore $O(|F| |tuples|)$, where $|F|$ is the number of features and $|tuples|$ is the number of transitions over which FIRM is computed. 

\begin{algorithm}[t] 
\caption{Compute feature importance for all features for given set of transitions.} \label{alg_firm}
\begin{algorithmic}[1]
  \Procedure{FIRM}{$tuples$} 
    \State $q_{tot} \leftarrow 0$ \Comment{expected value over full set}
    \For {$f \textrm{ in } \{1, \ldots, |F|\}$}
      \State $p_0[f] \leftarrow 0$ \Comment{ratio of set with $s[f] = 0$}
      \State $q_0[f] \leftarrow 0$ \Comment{$\mathbb{E}(s | s[f] = 0)$ for $s$ in set}
    \EndFor
    \For {$t \textrm{ in } tuples$}
      \State $g\_val \leftarrow g(t_s)$
      \State $q_{tot} \pluseq g\_val$ \Comment{store sum for $q_{tot}$}
      \For {$f \textrm{ in } \{1, \ldots, |F|\}$ }
        \If{$s[f] = 0$} \Comment{$p_0$ is tally, $q_0$ is sum}
          \State $p_0[f] \pluseq 1$
          \State $q_0[f] \pluseq g\_val$
        \EndIf
      \EndFor
    \EndFor
    \State $q_{tot} = q_{tot} / |tuples|$ \Comment{convert sum to average}
    \For { $f \textrm{ in } \{1, \ldots, |F|\}$} \Comment{intermediate terms}
      \State $q_0[f] \leftarrow q_0[f] / p_0[f]$
      \State $p_0[f] \leftarrow p_0[f] / |tuples|$
      \State $p_1[f] \leftarrow 1 - p_0$
      \State $q_1[f] \leftarrow (q_{tot} - p_0[f] q_0[f]) / p_1[f]$
      \State $q_{diff}[f] \leftarrow q_0[f] - q_1[f]$
    \EndFor
    \State $\textbf{return } [ q_{diff}[f] \sqrt{p_0[f] p_1[f]} 
    \textbf{ for } f \textrm{ in } \{1, \ldots, |F|\}]$
  \EndProcedure
\end{algorithmic}
\end{algorithm}

\subsubsection{APG Gen Runtime}
The runtime of Algorithm~\ref{alg_makePolicyGraph} is quadratic in the number of features and linear in the number of provided transitions, $O(|F|^2 |tr\_samples|)$.



Creating the initial abstract states (i.e., those based only on action taken) takes time $O(|A| + |tr\_samples|)$, where we assume $|A| \leq |tr\_samples|$. Computing FIRM for all of these abstract states takes $O(|F| |tr\_samples|)$ time. The while loop in lines 9 to 21 forms the bulk of the algorithm, which we will analyze last. Creating the lookup and transition tables takes $O(|tr\_samples|)$, assuming a zero matrix can be created in constant time for line 6. 

For lines 9 to 21, during each iteration of the while loop, the runtime is $O(\log_2(|c|))$ to insert the new $i_{max}$s if a max-heap is used to store the $j_{max}$s, $O(|c[i_{max}]|)$ to partition the set $c[i_{max}]$, and $O(|F| |c[i_{max}]|)$ to compute FIRM for both new sets. Note that each time a set is divided, the number of features within that set with non-fixed values (and therefore positive importance) is reduced by one. Therefore, any given tuple may only be part of an evaluated set $|F|$ times. As a result, over all iterations of the loop, the set division and FIRM computation takes at most $O(|F|^2 |tr\_samples|)$ time. This can happen over the course of up to $2^{|F|}$ divisions, so the max-heap insertion takes time at most $O(|F|)$. 

The overall worst-case runtime for APG Gen is then on the order of $O(|F|^2 |tr\_samples|)$. This is favorable since runtime must be at least linear in $|tr\_samples|$ and $F$ is generally small compared to the number of tuples. 

\section{Experimental Methodology}
We evaluate APG Gen on a novel domain with scalable state space and controllable stochasticity. We describe this domain, PrereqWorld, in Section~\ref{em_prereqworld}. Experimental settings are described in Section~\ref{em_settings}.

\subsection{PrereqWorld} \label{em_prereqworld}

We introduce the PrereqWorld domain for evaluating our approach. 
This domain is an abstraction of a production task where the agent is to create a specific, multi-component item using a number of manufacturing steps. 
The size of the state-space for an instance of this domain is controlled by the number of unique items, $m$. 
The agent may only have one of each item at a time.
Production of each item may require some prerequisites, a subset of the other items, but no cycle of dependencies is permitted. 
In producing an item, the prerequisite items are usually lost. 
A domain parameter, $\rho$, controls the probability that an item is lost. 

For ease of notation, we assume that the items are numbered according to their place in a topological sort (i.e., an item's prerequisites must be higher-numbered). 
Let $i_d$ refer to the desired final item. 
For each item $i_j$, let $C_j$ be the set of prerequisite items which $i_j$ requires. 
A sample MDP is shown in Figure~\ref{prereqworld_example_MDP}. 
Note how the goal is to make $i_1$ and it requires having $i_3$ and $i_4$. In turn, $i_3$ also requires $i_4$.

A state consists of $m$ binary features, where the binary feature $f_j$ corresponds to whether the agent has an item $i_j$. 
Any state where $s[f_d]=1$ is a terminal state. 
The distribution of initial states is uniform over all possible non-terminal states. 
The reward is $-1$ for transitioning to a non-terminal state and $0$ for transitioning to a terminal state. 
For simplicity, we take $\gamma$ to be $1$, but the optimal policies for any domain instance remain optimal for any $\gamma$ in the interval $(0,1]$.

There are $m$ actions where the action $a_j$ corresponds to attempting to produce item $i_j$. 
Actions for currently possessed items or for items with unmet prerequisites have no effect. 
That is, $P(s | s, a_j) = 1$ when feature $s[f_j] = 1$ or there is an $i_k \in C_j$ such that $s[f_k] = 0$. 
When an action is successful, $f_j$ is set to $1$ and each of item $i_j$'s prerequisites is used with probability $1-\rho$. That is, for all $i_k \in C_j$, $f_k$ is independently set to $0$ with probability $(1-\rho)$ and left as $1$ with probability $\rho$. 

For the MDP in Figure~\ref{prereqworld_example_MDP}, 
note that transitions are deterministic ($\rho = 0$) for simplicity and we do not show the transition function for states where $s[f_1] = 1$ ($i_1$ is present) since all such states are terminal. 
Notice how the domain can be solved optimally from the starting position $0000$ (no items present) using the action sequence $[a_4, a_3, a_4, a_1]$. 
This ensures that an $i_4$ is present before $i_3$ is made, and another $i_4$ is created as a prerequisite to creating $i_1$. 
This domain is suitable for explanation as it has inherent dependencies and sets of states which are treated identically.

\begin{figure}[t]
\begin{center}
\includegraphics[width=0.8\linewidth]{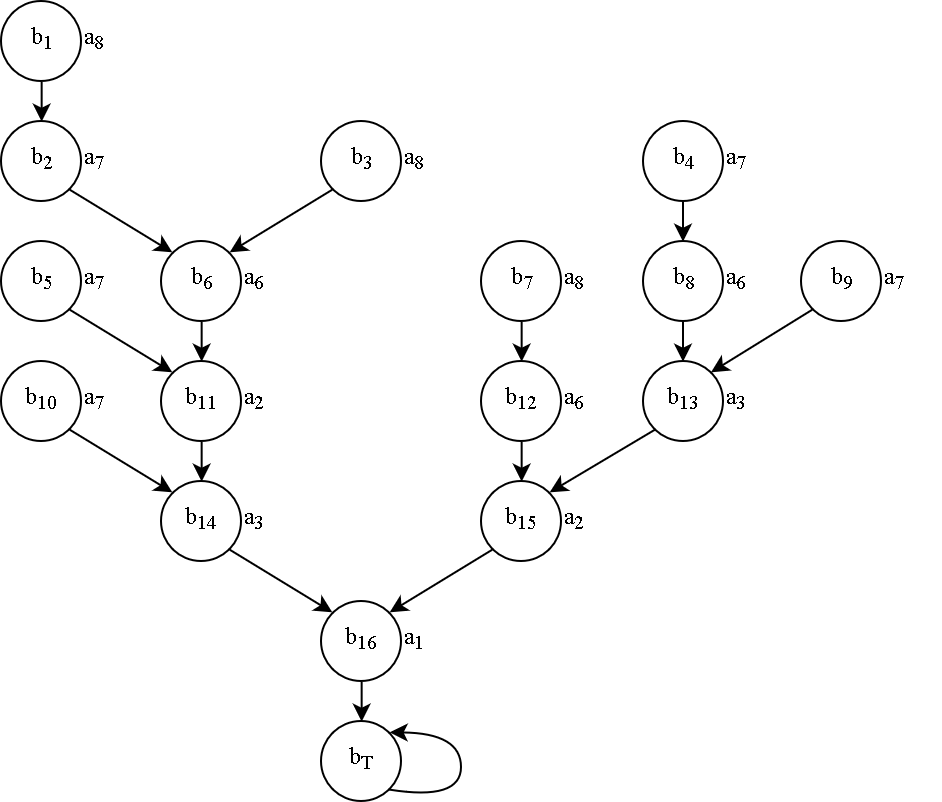}
\end{center}
\caption{An example APG made by APG Gen for a small PrereqWorld domain instance with $m=8$ and $\rho=0$. All edges have transition probability 1. The abstract state identifier is shown within each node, and the action taken is written adjacent to the node.}
\label{APG_example_prereq}
\end{figure}

An example APG made by APG Gen for an instance of PrereqWorld is given in Figure~\ref{APG_example_prereq}. APG Gen additionally describes each abstract state. For example, $b_{16}$ corresponds to all states where features 2 and 3 are 1. This corresponds to always taking action $a_1$ when an $i_2$ and $i_3$ are present, which corresponds to $C_1 = \{i_2, i_3\}$ in this domain instance. This correspondence between the domain constraints and the explanation would allow a human operator to verify that an agent is behaving as expected.

\begin{figure}[ht]
\begin{subfigure}[t]{0.5 \linewidth}
$m = 4$, $\rho = 0$, $i_d=i_1$ \\
$C_1 = \{i_3,i_4\}$\\
$C_2 = \{i_3,i_4\}$\\
$C_3 = \{i_4\}$\\
$C_4 = \{\}$\\
$S = \{0000, 0001, \ldots, 1111 \}$\\
$A = \{a_1, \ldots, a_4\}$\\
$T = \{1000, 1001, \ldots, 1111 \}$ \\
$R(s, a, s') = 0$ for $s' \in T$\\
$R(s, a, s') = -1$ for $s' \not \in T$\\
$\gamma = 1$
\end{subfigure}%
\begin{subfigure}[t]{0.5 \linewidth}
$P(0001 | 0000, a_4) = 1$\\
$P(0010 | 0001, a_3) = 1$\\
$P(0011 | 0010, a_4) = 1$\\
$P(1000 | 0011, a_1) = 1$\\
$P(0100 | 0011, a_2) = 1$\\
$P(0101 | 0100, a_4) = 1$\\
$P(0110 | 0101, a_3) = 1$\\
$P(0111 | 0110, a_4) = 1$\\
$P(1100 | 0111, a_1) = 1$\\
for other $s$ and $a$, \\
$P(s' | s, a) = 0$ when $s \neq s'$ \\
and \\
$P(s' | s, a) = 1$ when $s = s'$\\
\end{subfigure}
\caption{MDP for an example PrereqWorld instance.} \label{prereqworld_example_MDP}
\end{figure}

\subsection{Experimental Settings} \label{em_settings}

\subsubsection{APG Inputs}
For consistency, we use value iteration \cite{SuttonAndBarto} to create the policies and value functions used for experiments, but other methods could be used instead. 
We iterate until the state-value function no longer changes. 
To generate the transitions, we generated trajectories from a random starting state until the maximum number was reached. 

\subsubsection{APG Gen Stopping Criterion ($\epsilon$)}
In the case of binary features, FIRM corresponds to the expected change should the feature be changed from 0 to 1. 
Conveniently, sign is also preserved in the binary case, showing magnitude of importance as well as direction of effect. 
As a result, if no feature for any abstract state has FIRM magnitude greater than $\epsilon$, then changing any given feature is not expected to change the value of $g(s)$ by more than $\epsilon$ (e.g., $\mathbb{E}_{s \in c} ( | g(s, s_f=0) - g(s, s_f=1) | ) < \epsilon \forall c$). 
We use this as a guideline for setting $\epsilon$: we set $\epsilon$ to be the minimum difference in action-value between the best action and second-best action. 
For the PrereqWorld domain, this is $\epsilon = 1$.

\subsubsection{Trials}
For each plotted data-point, we generate 100 different PrereqWorld instances.
We evaluate each instance 1,000 times (i.e., we compute the feature importance for 1,000 different states or predict the $n$th action for 1,000 different trajectories), except for the points in Figure~\ref{fig:smallsize}, since the explanation size is fixed per APG.

\subsubsection{Domain Generation}
Each domain instance is parameterized by $\rho$ and $m$ as specified in Section~\ref{ex_results}. 
For simplicity, $d$ is always $i_1$. 
For each instance, we randomly add prerequisite relationships by selecting an item $i_j$ uniformly at random and then an item $i_k$ uniformly at random such that $k>j$. When adding prerequisite relationships, we constrain the expected number of actions to reach a terminal state to be within 10\% of $2m$.
This ensures that the domain can be solved in a reasonable amount of time using value iteration.

\section{Experimental Results} \label{ex_results}

\begin{figure}[t]
	\includegraphics[width=\linewidth]{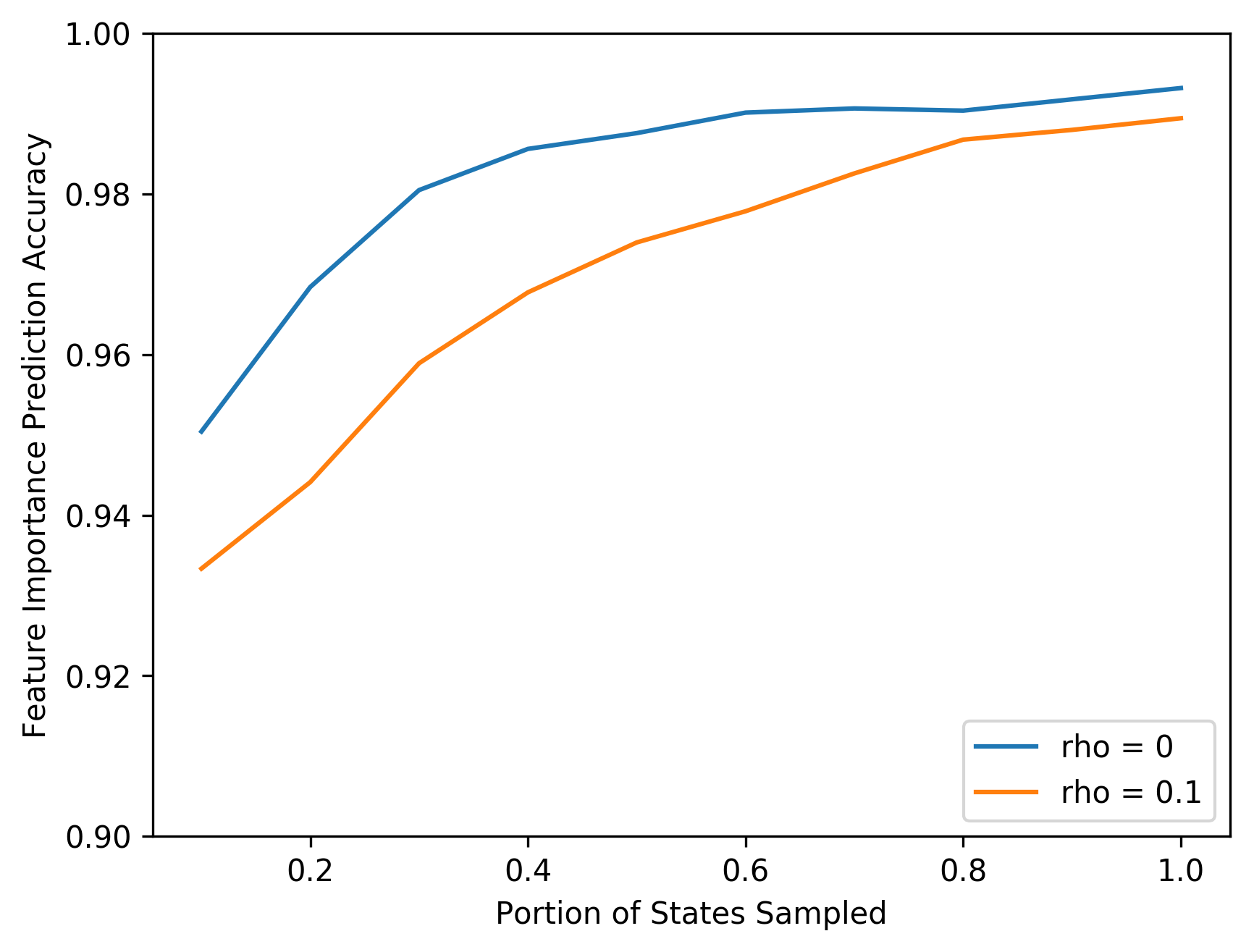}
	\caption{Comparison of feature importance prediction accuracy for increasing portion of non-terminal states.}
	\label{fig:generalize}
\end{figure}



\subsection{Local Explanation Generalization}
Based on the way we construct our abstract states, we can create ``if and only if'' conditions for a state in the transition sample set to be part of an abstract state's set, as described in Section~\ref{approach_abstractStateSummarization}. 
From this, we can create a local explanation consisting of the set of features which are important in that state. 
To evaluate how well APG Gen can generalize when predicting important features, we generate APGs using a set of transitions less than the full set of non-terminal states (i.e., we provide a set of transitions where no $(s, a, s')$ tuple shares an $s$ such that only a portion of non-terminal states appear as $s$). 
We then evaluate the local explanations by comparing to a ground truth computed for individual PrereqWorld instances with a domain parameter of $m=15$ ($|S| = 2^{15}$).

The portions of correct feature classifications (important vs. not important) are shown in Figure~\ref{fig:generalize}. 
APG Gen almost always correctly identifies the important features. 
Even with only 10\% of the states, the prediction is correct over 93\% of the time. 
When given 80\% of the states, predictions are correct 98.7\% of the time for both the stochastic and deterministic environments, which suggests that the model is able to identify genuine patterns in the policy. 
We believe the errors the system makes are caused by the splitting order induced by APG Gen's greedy splitting strategy. 

\begin{figure}[t]
	\includegraphics[width=\linewidth]{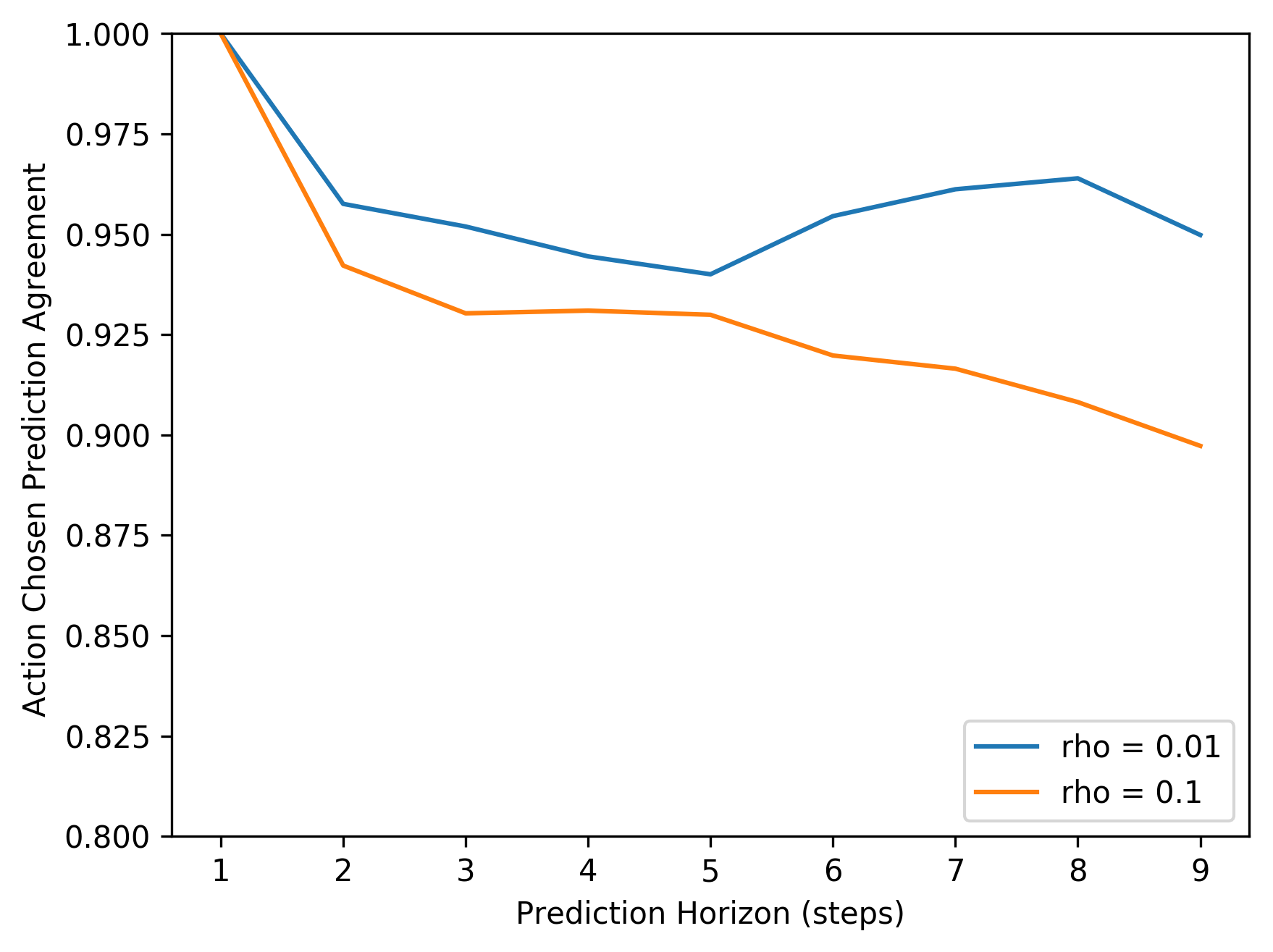}
	\caption{Action prediction for increasing time horizon.}
	\label{fig:markov}
\end{figure}

\subsection{$n$-hop Prediction Evaluation} \label{ex_markov_nhop}
An APG is able to predict the actions an agent will take, but this ability comes from an assumption made in Section~\ref{approach_graphRepresentation}. 
For each pair of abstract states, we produce a transition probability: the probability that the agent will be in the second abstract state, assuming the agent is following a transition tuple chosen at random from that first abstract state. 
This holds for a single action for states in the provided transition sample set, but not for arbitrary states and not when performing several of these predictions in sequence. 

To evaluate the error caused by making this assumption,
we have APG Gen predict the distribution of actions the agent will take $n$ time-steps in the future. 
We compare it to the true computed distribution and report the portion of actions for which the true and predicted distributions agree. 
This is for a domain parameter of $m=15$ ($|S| = 2^{15}|$). The size of the transition sample set is half the size of the set of non-terminal states. 

The action prediction is consistently correct when the domain is deterministic, so we report results for two stochastic domains in Figure~\ref{fig:markov}. 
Even with a small $\rho$, the prediction is less accurate as the number of steps increases, as is expected. 
However, there is no dramatic decrease, suggesting that the Markovian assumption made in Section~\ref{approach_graphConstruction} is reasonable. 
The steady decline is likely due to computing transition probabilities as an average of the transition sample set. 

\begin{figure}[t]
	\includegraphics[width=\linewidth]{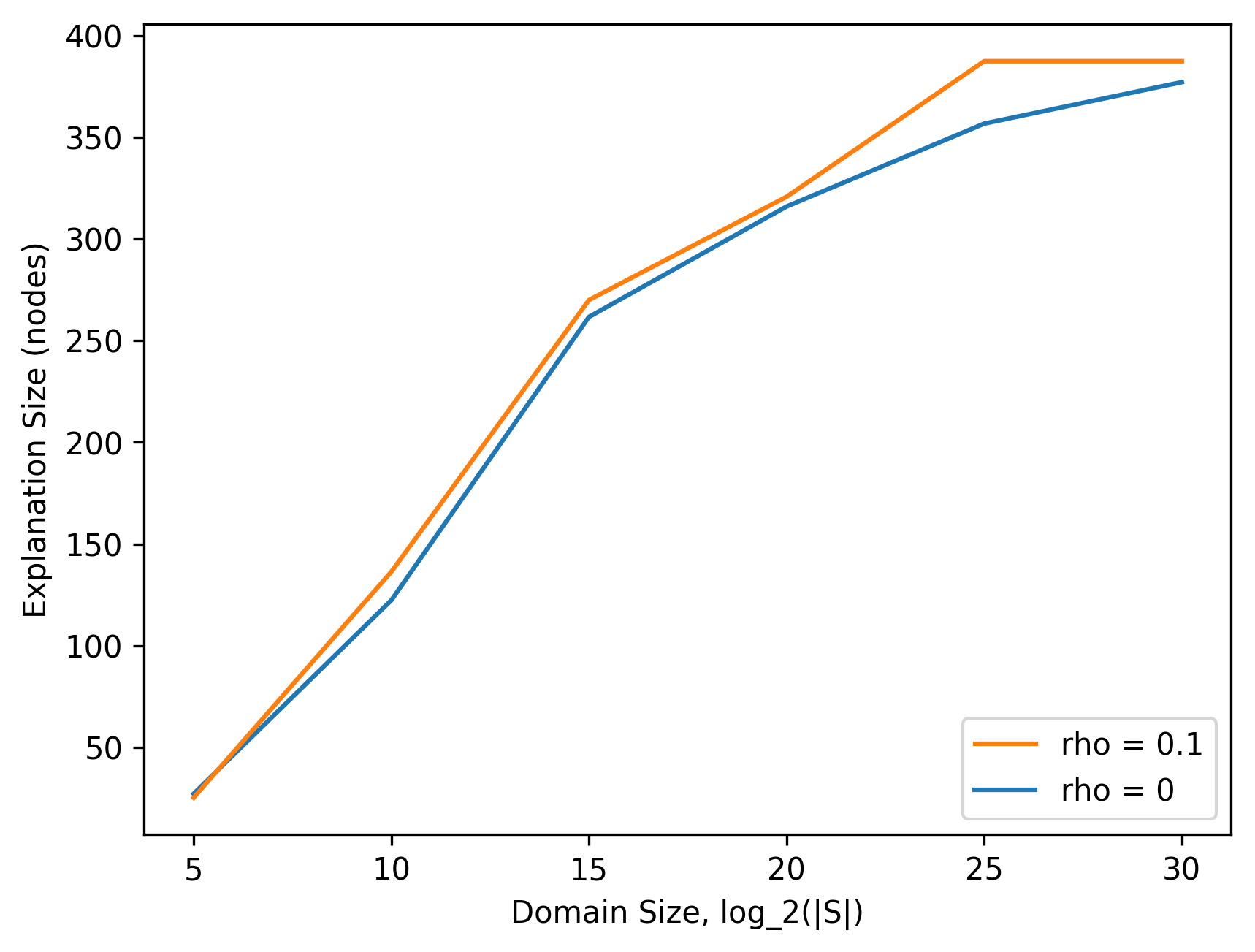}
	\caption{Comparison of explanation versus state-space size.}
	\label{fig:smallsize}
\end{figure}

\subsection{Explanation Size}
The purpose of APGs is to be more human-interpretable than a Markov chain made from the base MDP. 
Therefore, the number of nodes in an APG should be much lower than the number of grounded states in the base MDP. 
To test this, we construct domains with a number of states ranging from 32 to 1,073,741,824 and count the number of abstract states in the corresponding APG. 
As in Section~\ref{ex_markov_nhop}, for each generated APG, the size of the transition sample set is half the size of the set of non-terminal states. 
The results are presented in Figure~\ref{fig:smallsize}. 
Note that the x-axis is in log-scale. 

The explanation size grows sub-linearly in $m$ while the state-space size grows exponentially in $m$. 
This suggests that the explanation size is based more on the number of actions required to reach a terminal state than the number of states, which indicates that compact policy representations are being automatically extracted. 

\section{Conclusion and Future Work}
We introduced Abstract Policy Graphs, a whole-policy explanation from which state-specific explanations can be extracted. In addition, we presented APG Gen, an algorithm for creating an APG given a policy, learned value function, and set of transitions, without constraints on how these are created. We showed that APG Gen runs in time quadratic in the number of features and linear in the number of transitions provided, $O(|F|^2 |tr\_samples|)$. Additionally, we demonstrated empirical results showing the small size of the APGs relative to the original MDPs, as well as the types and quality of explanations which can be extracted. Together, these show that APG Gen can produce concise policy-level explanations in a tractable amount of time. 
Future work includes restructuring the explanations extracted from an APG to be better understood by a non-expert.
To address this, we are in the process of conducting a user study to evaluate the usefulness of APG explanations in different presentation formats. 

\section{Acknowledgements}
This material is based upon work supported by DARPA grants FA87501720152 and FA87501620042. Any opinions, findings and conclusions, or recommendations expressed in this material are those of the authors and do not necessarily reflect the views of DARPA.

\bibliography{main.bib}
\bibliographystyle{aaai}
\end{document}